\documentclass[letterpaper]{article} 
\usepackage{aaai2026}  

\usepackage{times}  
\usepackage{helvet}  
\usepackage{courier}  
\usepackage[hyphens]{url}  
\usepackage{graphicx} 
\usepackage{amssymb} 
\usepackage{natbib}  
\usepackage{caption} 
\usepackage{amsmath}
\usepackage{algorithm}
\usepackage{algorithmic}

\usepackage{newfloat}
\usepackage{listings}
\DeclareCaptionStyle{ruled}{labelfont=normalfont,labelsep=colon,strut=off} 
\floatstyle{ruled}
\newfloat{listing}{tb}{lst}{}
\floatname{listing}{Listing}
\title{WGDnet: Wishart-guided Geometric-aware Deep Network for PolSAR Image Classification}
\author{
     Junfei Shi,
Haojia Zhang,
Yu Cheng,
Yuke Li
}
    
\affiliations{

}

\usepackage{bibentry}

\begin{document}

\maketitle

\begin{abstract}
Polarimetric Synthetic Aperture Radar (PolSAR) classification underpins all-weather Earth observation. Conventional Wishart methods depend on rigid handcrafted operators with limited adaptability, while mainstream deep networks ignore PolSAR’s native Wishart scattering statistics. Additionally, fixed convolution windows fail to capture multi-scale, multi-directional terrain patterns, harming boundary detection and small-object characterization. To mitigate these drawbacks, we propose WGDNet, a Wishart-guided geometric-aware deep network. It integrates three core designs: (1) learnable Wishart convolutions with directional kernels for multi-scale statistical edge feature extraction; (2) an orientation-prior aggregation module that estimates dominant local directions and confidences to refine directional Wishart outputs adaptively; (3) GAnet, a scale-direction adaptive geometric-aware convolution that dynamically reshapes sampling grids to model anisotropic terrain and retain fine details. Our contributions lie in learnable Wishart statistical modeling, orientation-prior feature aggregation, and geometry-adaptive convolution. Evaluations across four real PolSAR datasets verify WGDNet surpasses existing state-of-the-art approaches in classification accuracy and boundary fidelity.

Keywords: PolSAR classification; learned Wishart convolution; directional prior; geometric-aware convolution; Scale-direction adaptive\textbf{}
\end{abstract}


\section{Introduction}

Polarimetric Synthetic Aperture Radar (PolSAR) provides full-polarization backscattering measurements that characterize the intrinsic scattering properties of ground targets\cite{551935}. Pixel-level PolSAR image classification is therefore fundamental to a wide range of Earth observation applications, including land-cover mapping, crop monitoring, disaster assessment\cite{7762055}, and urban planning. Unlike optical remote sensing, PolSAR can operate under all-day and all-weather conditions, enabling reliable terrain interpretation in challenging environments. Consequently, \cite{10530067}developing accurate and robust PolSAR classification methods remains an important topic in remote sensing and artificial intelligence\cite{8517524}.

Traditional PolSAR classification methods are mainly based on physically motivated statistical models\cite{Yang_Sun_Duan_Cheng_2025}. Representative approaches include Wishart maximum-likelihood classification, Wishart edge detection\cite{267879}, polarimetric decomposition methods such as Pauli, Freeman--Durden\cite{10281574}, and Yamaguchi decompositions\cite{1487628}, and statistical clustering algorithms combining K-means with Wishart likelihood metrics. These methods exploit the statistical characteristics of PolSAR covariance or coherency matrices and provide physically interpretable scattering representations. However, they generally rely on handcrafted operators, fixed-size sliding windows, and designed classifier\cite{rs11242994}. Such designs lack sufficient adaptability to heterogeneous scenes and cannot be optimized jointly with the final classification objective, limiting their ability to characterize multi-scale and arbitrarily oriented terrain structures\cite{5687997}.

With the rapid development of deep learning, data-driven neural networks have been widely applied to PolSAR image classification\cite{10097620}. Convolutional neural networks, Vision Transformers\cite{9658539}, and hybrid architectures\cite{10488702} can automatically learn high-level semantic representations from PolSAR data and have achieved substantial performance improvements. Nevertheless, existing deep PolSAR classification models still face two major limitations.

1) Conventional deep learning pipelines flatten PolSAR matrices into 9D input vectors, discarding the covariance matrix’s scattering statistics and intrinsic geometric properties\cite{10599283}\cite{8809406}. The resulting features suffer severe speckle interference and weak physical consistency, hurting generalization across sensors, scenes, and land-cover categories\cite{rs12040655}.

2) Standard convolutions adopt fixed square kernels with symmetric horizontal and vertical receptive fields\cite{10.1609/aaai.v38i7.28502}. Yet PolSAR imagery contains anisotropic, arbitrarily shaped targets (small vegetation, elongated corridors, large farmlands, irregular terrain edges). Fixed-shape convolutions fail to capture these structures,\cite{Yuan_Zheng_Li_Liu_Liu_Li_Hou_Cheng_2026} and recurrent downsampling erases fine details to yield blurry classification boundaries\cite{rs14153737}.

To jointly exploit polarimetric scattering statistics and adaptive geometric information\cite{7920279}\cite{8292839}, we propose a \textbf{Wishart-guided Geometric-aware Deep Network}, termed \textbf{WGDNet}, for PolSAR image classification. WGDNet integrates Wishart statistical modeling, orientation-prior-guided response aggregation, and scale--direction adaptive geometric feature extraction into a unified end-to-end framework.

Specifically, our framework comprises three core components. First, a learnable Wishart convolution branch embeds Wishart likelihood-ratio statistics into the network. Trainable soft-split local kernels estimate multi-scale, multi-orientation statistical disparities to produce PolSAR-statistics-compliant task-specific edge features. Second, the proposed OriPriorNet predicts each pixel’s dominant local orientation and confidence. It generates orientation prior scores via periodic angular distances between predicted orientations and preset directional candidates to adaptively aggregate multi-directional Wishart outputs, amplifying terrain-aligned edge responses and suppressing noisy directional signals. Third, we present GAnet, a scale-direction adaptive geometric-aware convolution module. It dynamically predicts per-pixel sampling dimensions and rotation to build rotated rectangular sampling grids for anisotropic terrain modeling, boosting feature representation for elongated targets, irregular zones, and fine-grained objects.

The main contributions of this work are summarized as follows:

\begin{itemize}
    \item We propose a learnable Wishart statistical modeling branch that incorporates the physical distribution prior of PolSAR data into deep networks through trainable soft-split kernels. The resulting multi-scale and multi-orientation statistical responses can be optimized jointly with the classification objective.

    \item We develop an orientation-prior-guided aggregation strategy. OriPriorNet estimates pixel-wise dominant orientations and confidence values, while a periodic angular distance metric is used to generate directional prior weights for adaptive fusion of Wishart responses.

    \item We design a scale--direction adaptive GAnet that predicts pixel-wise geometric parameters and constructs rotated rectangular sampling grids. This module improves the modeling of anisotropic terrain structures and alleviates the loss of small-target and boundary information.

    \item Extensive experiments on four real PolSAR datasets demonstrate that WGDNet consistently outperforms representative physically driven and deep learning-based methods, achieving higher classification accuracy and more complete boundary preservation.
\end{itemize}




\section{Relate Work}
\subsection{Deep Learning-based Classification}


Traditional PolSAR classification relies on target decomposition~\cite{673687} and statistical models~\cite{8517360} (e.g., complex Wishart distribution~\cite{1183689}) to analyze covariance matrices and characterize polarimetric scattering properties~\cite{9165101}. Though physically interpretable, these handcrafted methods suffer from limited generalization in complex scenarios. Recent deep learning advances, including CNNs, complex-valued networks, attention mechanisms, and Transformers, have substantially improved PolSAR feature representation~\cite{rs16101676}. For instance, Zhang et al. proposed a complex-valued convolutional neural network~\cite{8039431} to preserve the phase and amplitude information of polarimetric scattering parameters. Nevertheless, most existing deep learning approaches flatten PolSAR inputs into feature vectors and ignore the intrinsic statistical characteristics of covariance matrices~\cite{9424197}, failing to fully preserve physical scattering priors during optimization. This motivates the development of learnable frameworks that synergize physical scattering constraints with data-driven feature learning for robust PolSAR classification.

\subsection{Scale-adaptive Learning Methods}

Standard CNN\cite{10.1007/s11263-021-01501-8} adopt fixed sampling grids, limiting their ability to model objects with variable scales, orientations, and irregular geometric structures . To alleviate this limitation, scale-adaptive networks have been extensively explored. Multi-scale and large-kernel architectures expand receptive fields to aggregate multi-range contextual cues but are restricted by predefined kernel configurations and static sampling layouts\cite{article},\cite{9880273}. Deformable convolutions\cite{8237351} improve spatial modeling flexibility via learnable sampling offsets but fail to explicitly encode object scale and orientation priors . Rotation-equivariant and oriented convolutions boost directional feature extraction yet remain constrained by fixed orientation templates and scale-invariant receptive fields\cite{8100010,han2021redetrotationequivariantdetectoraerial}. Recently, ARConv\cite{11093597} enables dynamic kernel dimension adjustment to accommodate objects with diverse scales and aspect ratios . Nevertheless, it lacks explicit local orientation estimation and cannot fully excavate the inherent directional scattering signatures of PolSAR data. In essence, current approaches struggle with the joint adaptive modeling of sampling scale and orientation, resulting in insufficient characterization of anisotropic terrain structures and blurry boundary discrimination in PolSAR classification tasks.

\begin{figure*}[t]
	\centering
    \includegraphics[scale=0.36]{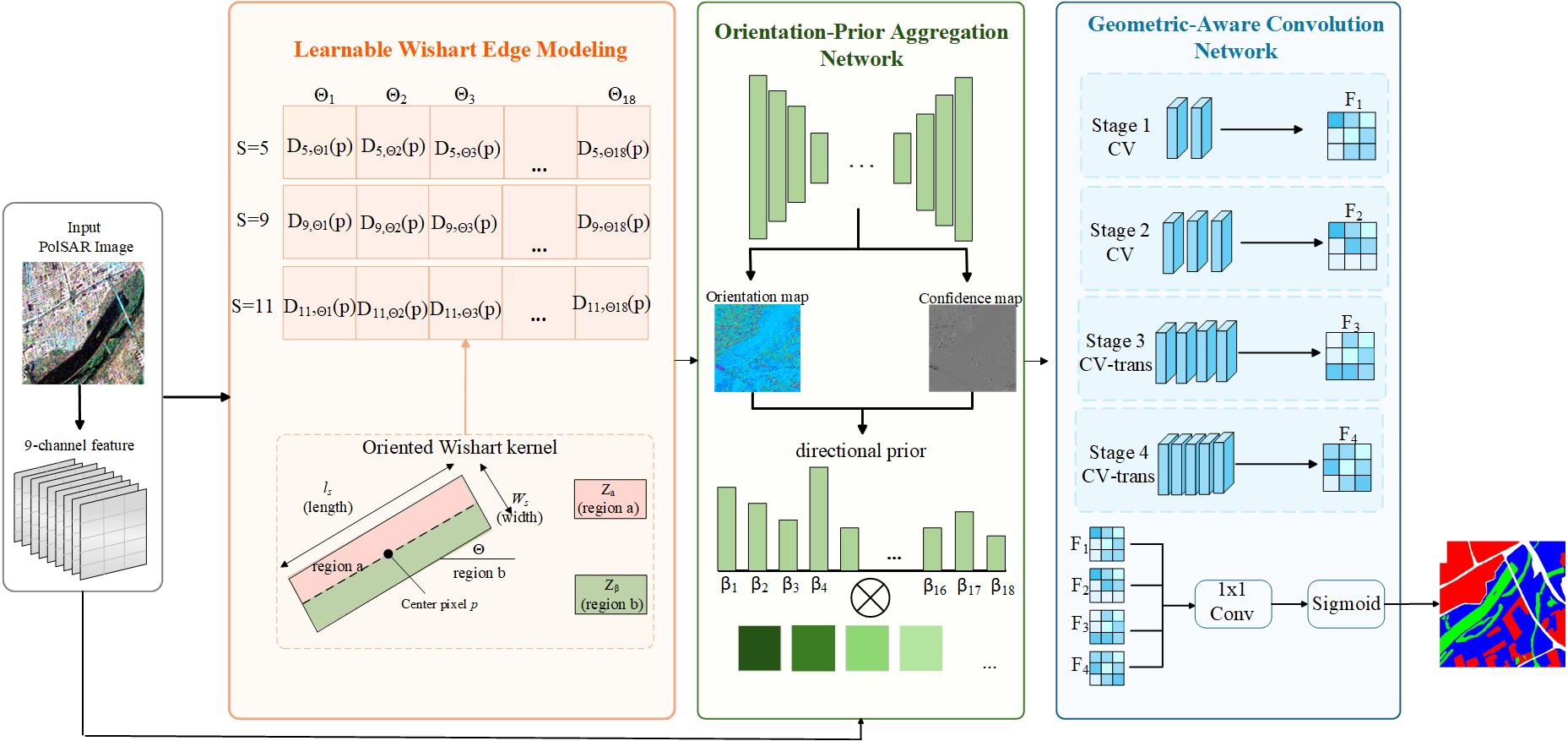}
	\caption{Framework of the proposed wishart-guided geometric-aware
deep network for PolSAR image classification.}
	\label{fig2}
\end{figure*}
\section{Proposed Method}
This section presents a Wishart-guided Geometric-aware Deep Network (WGDNet) for PolSAR image classification, whose framework is illustrated in Fig.~\ref{fig2}. The proposed method integrates learnable Wishart statistical edge modeling, direction-prior-guided multi-orientation aggregation, and direction-aware feature representation. Specifically, we design learnable Wishart edge convolution operators to capture multi-scale and multi-directional features. We then introduce a primary direction-guided multi-orientation aggregation module to adaptively fuse multi-directional features. Furthermore, scale-direction aware convolution kernels are developed to adaptively model diverse terrain characteristics, in which a scale-direction adaptive convolution window is designed in place of the fixed-square window used in standard convolutions.
\subsection{Learnable Wishart Edge Modeling}
\label{subsec:wishart_edge}

Given a PolSAR image, its input feature is denoted by
$\mathbf{X}\in\mathbb{R}^{H\times W\times 9}$.
For each pixel position $p$, the nine-dimensional polarimetric feature is reorganized into a $3\times3$ Hermitian coherency matrix $\mathbf{T}_{p}$:
\begin{equation}
\mathbf{T}_{p}=
\begin{bmatrix}
\left\langle |S_{hh}|^{2}\right\rangle &
\left\langle S_{hh}S_{hv}^{*}\right\rangle &
\left\langle S_{hh}S_{vv}^{*}\right\rangle \\
\left\langle S_{hv}S_{hh}^{*}\right\rangle &
\left\langle |S_{hv}|^{2}\right\rangle &
\left\langle S_{hv}S_{vv}^{*}\right\rangle \\
\left\langle S_{vv}S_{hh}^{*}\right\rangle &
\left\langle S_{vv}S_{hv}^{*}\right\rangle &
\left\langle |S_{vv}|^{2}\right\rangle
\end{bmatrix}.
\label{eq:coherency_matrix}
\end{equation}
where $S_{hh}$ denotes the complex scattering echo with horizontal transmitting and horizontal receiving mode. The superscript $*$ denotes complex conjugation, and $\langle\cdot\rangle$ represents ensemble averaging.

Conventional deep learning approaches flatten PolSAR covariance matrices into an input vector, which ignores both the PolSAR data's statistical characteristics and geometric structure of complex matrices. Since PolSAR covariance matrices obey the Wishart distribution, we define learnable Wishart convolution kernels to capture multi-scale, multi-orientation PolSAR scattering signatures.

Specifically, for each scale $s\in\mathcal{S}$ and orientation $\theta\in\Theta$, a local statistical kernel is softly divided into two complementary regions, denoted by $a$ and $b$. For each region $r\in\{a,b\}$, the locally weighted mean coherency matrix is calculated as
\begin{equation}
\overline{\mathbf{T}}_{r}(p)=
\frac{1}{Z_{r}(p)}
\sum_{p_i\in\mathcal{R}_{r}(p)}
w_{p_i}^{(s,\theta,r)}\mathbf{T}_{p_i},
\label{eq:weighted_mean_matrix}
\end{equation}
where $\mathcal{R}_{r}(p)$ denotes the pixels belonging to region $r$ within the local window centered at $p$, $w_{p_i}^{(s,\theta,r)}$ is the corresponding learnable kernel weight, and
\begin{equation}
Z_{r}(p)=
\sum_{p_i\in\mathcal{R}_{r}(p)}
w_{p_i}^{(s,\theta,r)}
\end{equation}
is the normalization term. These kernel weights are optimized end-to-end using the final classification objective.



Based on the weighted mean coherency matrices
$\overline{\mathbf{T}}_a(p)$ and
$\overline{\mathbf{T}}_b(p)$, the Wishart likelihood ratio
at scale $s$ and orientation $\theta$ is defined as
\begin{equation}
\begin{aligned}
Q_{s,\theta}(p)
=
\Bigg[
\frac{
\left|\overline{\mathbf{T}}_a(p)\right|^{N_a(p)}
\left|\overline{\mathbf{T}}_b(p)\right|^{N_b(p)}
}{
\left|
\frac{
N_a(p)\overline{\mathbf{T}}_a(p)
+N_b(p)\overline{\mathbf{T}}_b(p)
}{
N_a(p)+N_b(p)
}
\right|^{N_a(p)+N_b(p)}
}
\Bigg]^L .
\end{aligned}
\label{eq:wishart_q}
\end{equation}

The corresponding statistical response is defined as
\begin{equation}
D_{s,\theta}(p)
=
-2\rho\log Q_{s,\theta}(p),
\label{eq:wishart_response}
\end{equation}
where $L$ denotes the number of equivalent looks and
$\rho$ is the correction coefficient.

Repeating this procedure over all scales and orientations yields the multi-scale and multi-orientation Wishart response set:
\begin{equation}
\mathcal{D}(p)=
\left\{
D_{s,\theta}(p)
\mid
s\in\mathcal{S},\,
\theta\in\Theta
\right\}.
\label{eq:wishart_response_set}
\end{equation}
Different scales capture statistical variations over different spatial ranges, while different orientations characterize boundary structures along different directions. These responses are subsequently organized as Wishart statistical edge features and passed to the following feature aggregation module.

\subsection{Orientation-Prior Aggregation Network}
\label{subsec:orientation_prior}

Wishart statistical responses encode multi-scale and multi-orientation cues, yet distinct orientations exert unequal impacts on pixel discrimination. For areas with prominent directional terrain structures, responses consistent with local dominant orientations carry more discriminative edge information. To adaptively aggregate multi-orientation Wishart features, we incorporate orientation priors to weight meaningful directional responses.

In this paper, an orientation-prior network, termed OriPriorNet, is designed to estimate the dominant orientation and its confidence from the input PolSAR features:
\begin{equation}
\left(\phi_p,c_p\right)
=
\operatorname{OriPriorNet}(\mathbf{X})_p,
\label{eq:ori_prior}
\end{equation}
where $\phi_p$ denotes the estimated dominant orientation around pixel $p$, and $c_p\in[0,1]$ represents the confidence of the orientation estimation. Since boundary orientations exhibit $\pi$-periodicity, i.e., $\phi$ and $\phi+\pi$ describe the same structural direction, the orientation is restricted to $\phi_p\in[0,\pi)$.

Let the discrete orientation set be
$\Theta=\{\theta_1,\theta_2,\ldots,\theta_K\}$.
The periodic angular distance between the predicted orientation $\phi_p$ and the $k$-th candidate orientation $\theta_k$ is defined as
\begin{equation}
\delta_k(p)
=
\min\left(
\left|\phi_p-\theta_k\right|,
\pi-\left|\phi_p-\theta_k\right|
\right).
\label{eq:periodic_distance}
\end{equation}
This formulation avoids treating orientations close to $0$ and $\pi$ as distant directions.

Based on the angular distance, the corresponding orientation-prior score is calculated as
\begin{equation}
\alpha_k(p)
=
\exp\left(
-\frac{\delta_k^2(p)}{2\sigma^2}
\right),
\label{eq:orientation_prior_score}
\end{equation}
where $\sigma$ controls the smoothness of the orientation prior. A smaller angular distance produces a larger prior score.

To reduce the influence of inaccurate orientation estimates in structurally ambiguous regions, the prior score is further modulated by the confidence map:
\begin{equation}
\widetilde{\alpha}_k(p)
=
c_p\alpha_k(p)
+
\left(1-c_p\right)\frac{1}{K}.
\label{eq:confidence_prior}
\end{equation}
When the local orientation is reliable, responses aligned with the predicted direction are emphasized. Otherwise, the prior approaches a uniform distribution, allowing the aggregation to rely mainly on data-driven responses.

For the response at scale $s$ and orientation $\theta_k$, a data-driven attention score $A_{s,k}(p)$ is first generated from the Wishart response $D_{s,\theta_k}(p)$. The orientation prior is then incorporated as
\begin{equation}
E_{s,k}(p)
=
A_{s,k}(p)
+
\lambda\widetilde{\alpha}_k(p),
\label{eq:prior_guided_score}
\end{equation}
where $\lambda$ controls the contribution of the orientation prior. The normalized orientation attention weight is obtained by
\begin{equation}
\beta_{s,k}(p)
=
\frac{
\exp\left(E_{s,k}(p)\right)
}{
\displaystyle\sum_{k'=1}^{K}
\exp\left(E_{s,k'}(p)\right)
}.
\label{eq:orientation_attention}
\end{equation}

Finally, the orientation-aggregated Wishart response at scale $s$ is formulated as
\begin{equation}
D_s(p)
=
\sum_{k=1}^{K}
\beta_{s,k}(p)
D_{s,\theta_k}(p).
\label{eq:orientation_aggregation}
\end{equation}
Unlike average pooling or max pooling over the orientation dimension, the proposed aggregation adaptively emphasizes statistically informative responses according to the local structural orientation. The resulting multi-scale responses are subsequently used as direction-aware statistical priors for deep feature extraction.

\subsection{Geometric-Aware Convolution Network}
\label{subsec:cvtrans}

After obtaining the orientation-guided multi-scale Wishart responses $\{D_s(p)\}$, they are fused with the original PolSAR features and passed to the pixel-wise classification backbone. The input feature of the $l$-th layer can be expressed as
\begin{equation}
\mathbf{F}_l
=
\operatorname{Fusion}
\left(
\mathbf{X},
\{D_s\}_{s\in\mathcal{S}}
\right).
\label{eq:feature_fusion}
\end{equation}

A conventional convolution uses a fixed square sampling grid and can be written as
\begin{equation}
\mathbf{F}_{l+1}
=
\sigma\left(
\mathbf{W}*\mathbf{F}_l+\mathbf{b}
\right),
\label{eq:standard_conv}
\end{equation}
where $*$ denotes convolution, $\mathbf{W}$ and $\mathbf{b}$ are the convolution weights and bias, respectively, and $\sigma(\cdot)$ denotes the nonlinear activation function.

\begin{figure}[!t]
	\centering
    \includegraphics[scale=0.36]{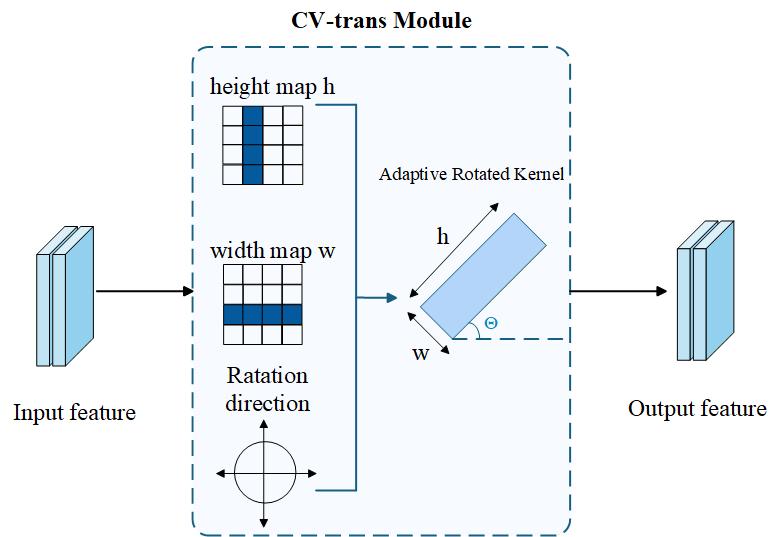}
	\caption{Structure of the proposed CV-Trans module.}
	\label{fig3}
\end{figure}

However, fixed-square convolution kernel structures are insufficient for modeling elongated targets, oblique boundaries, and irregular terrain regions. To address this limitation, the GAnet module design a geometric-aware convolution kernel, which predicts a height map $h(p)$, a width map $w(p)$, and an orientation map $\theta(p)$ for each spatial position. 
These geometric parameters enable the convolution kernel to adapt its scale and orientation to local structures. The height and width maps determine the sampling region, while the orientation map controls kernel rotation, transforming the fixed grid into a scale- and direction-adaptive sampling pattern, as illustrated in Fig.~\ref{fig3}

\textbf{GA convolution kernel:} For the $(i,j)$-th point of a rectangular sampling kernel, the scale-adaptive offset is defined as
\begin{equation}
\mathbf{r}_{i,j}(p)
=
\begin{bmatrix}
i\,h(p)\\
j\,w(p)
\end{bmatrix}.
\label{eq:scaled_offset}
\end{equation}
The offset is then rotated according to the predicted local orientation:
\begin{equation}
\mathbf{q}_{i,j}(p)
=
\mathbf{p}
+
\begin{bmatrix}
\cos\theta(p) & -\sin\theta(p)\\
\sin\theta(p) & \cos\theta(p)
\end{bmatrix}
\mathbf{r}_{i,j}(p),
\label{eq:rotated_sampling}
\end{equation}
where $\mathbf{p}$ denotes the coordinate of the central pixel, and $\mathbf{q}_{i,j}(p)$ is the transformed sampling position. The indices $i$ and $j$ correspond to the vertical and horizontal sampling positions of the rectangular kernel, respectively.

Since the transformed coordinates are generally non-integer, bilinear interpolation is employed to obtain feature values from the four neighboring grid points. In this way, the original fixed sampling grid is converted into a scale- and orientation-adaptive rectangular structure.

With the GA convolution kernel, the output of the GAnet module is formulated as
\begin{equation}
\mathbf{F}_{l+1}
=
\mathcal{C}_{H_k,W_k}
\left(
\mathcal{S}
\left(
\mathbf{F}_l;
h,w,\theta
\right)
\right)
\odot
\mathcal{M}(\mathbf{F}_l)
+
\mathcal{B}(\mathbf{F}_l),
\label{eq:cvtrans}
\end{equation}
where $\mathcal{S}(\cdot)$ denotes adaptive scale--direction sampling, $\mathcal{C}_{H_k,W_k}(\cdot)$ represents rectangular convolution aggregation with $H_k\times W_k$ sampling points, and $\mathcal{M}(\cdot)$ and $\mathcal{B}(\cdot)$ denote the modulation and bias branches, respectively. The symbol $\odot$ denotes element-wise multiplication.

This adaptive mechanism enables GAnet to dynamically adjust its receptive-field shape and orientation according to local terrain geometry. Standard convolutions are employed in shallow stages to preserve local details, whereas GAnet is introduced in deeper stages to capture direction-aware contextual information. Multi-level features are progressively fused to generate the final pixel-wise representation $\mathbf{F}$.

\subsection{Prediction and Optimization Objective}
\label{subsec:prediction_loss}

Given the final fused feature $\mathbf{F}$, a lightweight classification head $\Psi(\cdot)$ maps each pixel feature into the semantic class space. For pixel $p$, the class probability vector is defined as
\begin{equation}
\widehat{\mathbf{y}}_p
=
\Psi(\mathbf{F}_p)
=
\operatorname{Softmax}
\left(
\mathbf{W}_c\mathbf{F}_p+\mathbf{b}_c
\right),
\label{eq:classification_probability}
\end{equation}
where $\mathbf{F}_p$ denotes the feature representation of pixel $p$, $\mathbf{W}_c$ and $\mathbf{b}_c$ are the parameters of the classification head, and
$\widehat{y}_{p,c}\in[0,1]$ represents the predicted probability that pixel $p$ belongs to class $c$.

The final classification result is determined by
\begin{equation}
\widehat{c}_p
=
\underset{c\in\{1,\ldots,C\}}{\arg\max}
\ \widehat{y}_{p,c},
\label{eq:final_prediction}
\end{equation}
where $C$ denotes the number of land-cover classes.

During training, only labeled foreground pixels are used for supervision. Let $\mathcal{P}_{\mathrm{lab}}$ denote the set of labeled training pixels and $y_{p,c}$ denote the corresponding one-hot label. The class-weighted cross-entropy loss is defined as
\begin{equation}
\mathcal{L}_{\mathrm{cls}}
=
-\frac{1}{|\mathcal{P}_{\mathrm{lab}}|}
\sum_{p\in\mathcal{P}_{\mathrm{lab}}}
\sum_{c=1}^{C}
\omega_c
y_{p,c}
\log\left(
\widehat{y}_{p,c}
\right),
\label{eq:classification_loss}
\end{equation}
where $\omega_c$ denotes the weight assigned to class $c$ to alleviate class imbalance.

The entire network is trained end-to-end by minimizing $\mathcal{L}_{\mathrm{cls}}$. Therefore, the learnable Wishart statistical modeling, orientation-prior estimation, multi-orientation response aggregation, and GAnet feature extraction modules are jointly optimized under the final pixel-wise classification objective.

\section{Experiments}
\subsection{Experimental Data and Settings}
We evaluate the proposed WGDNet on three widely used PolSAR benchmark datasets, namely Xi'an, San Francisco, Flevoland1 and Flevoland2. Detailed descriptions of the datasets, preprocessing procedures, and visualization examples are provided in the supplementary material.

The proposed model is implemented in PyTorch and trained using the Adam optimizer with an initial learning rate of (\(1\times10^{-3}\)) for 300 epochs. Following common PolSAR classification protocols, (5\%) of the labeled pixels from each class are randomly selected for training, (1\%) for validation, and the remainder for testing. Overall Accuracy (OA), Average Accuracy (AA), and the Kappa coefficient are adopted as quantitative metrics.We compare WGDNet with six representative methods, including PolMPCNN (PMC), HybridCVNet (HCV), SpectralDiff (SDiff), NGDiff-SM (NGDiff), SDNet (SD), and F-Conv (FConv). Detailed descriptions and implementation settings are provided in the supplementary material.

\subsection{Experimental Result and Analysis}
\subsubsection{Xi'an data set}
\begin{figure}[!t]
	\centering
    \includegraphics[scale=0.4]{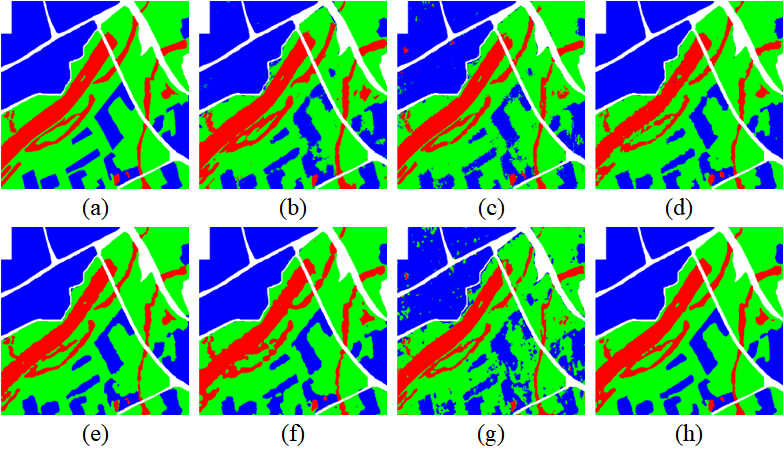}
	\caption{Classification maps of the Xi'an data set. (a) Ground truth; (b) PolMPCNN; (c) HybridCVNet; (d) SpectralDiff; (e) NGDiff-SM; (f) SDNet; (g) F-Conv; (h) Proposed.}
	\label{fig4}
\end{figure}

As shown in Figs.~\ref{fig4}(b)--(h), the compared methods generally recover the main land-cover regions, but still exhibit scattered errors, fragmented predictions, detail loss, boundary confusion, or speckle noise. In contrast, WGDNet produces more homogeneous intra-class regions and clearer class boundaries by jointly exploiting Wishart statistical cues and adaptive geometric modeling. As reported in Table~\ref{tab:xian_results}, WGDNet achieves the best OA, AA, and Kappa values of (97.94\%), (97.73\%), and (96.61\%), respectively, while also obtaining the highest accuracy for all three classes, demonstrating more balanced and reliable classification performance.

\begin{table}[t]
\centering
\caption{Classification results (\%) of different methods on the Xi'an dataset.}
\label{tab:xian_results}
\resizebox{\columnwidth}{!}{
\begin{tabular}{lccccccc}
\hline
Class & PMC & HCV & SPD & NGD & SDN & FC & Ours\\
\hline
Water
& \underline{95.52}
& 92.01
& 89.70
& 92.01
& 92.02
& 91.75
& \textbf{96.57} \\

Grass
& 90.95 
& 94.99 
& 97.09
& \underline{97.43}
&95.13
& 89.39
& \textbf{97.86} \\

Building
& 97.68
& 95.05 
& 97.74
&  \underline{98.43}
& 98.16
& 88.46
& \textbf{98.79} \\
\hline
OA
& 94.01 
& 94.56
& 96.21
& \underline{96.99}
& 95.98
& 89.38
& \textbf{97.94} \\

AA
& 94.71
& 94.02
&  94.85
&  \underline{95.98}
& 95.65
& 88.16
& \textbf{97.73} \\

Kappa
& 90.25
& 91.02
&  93.73
& 82.75
& \underline{93.40}
& 82.37
& \textbf{96.61} \\
\hline
\end{tabular}
}
\end{table}

\subsubsection{ San Francisco data set}As shown in Figs.~\ref{fig5}(b)--(h), the compared methods exhibit different levels of spatial consistency and boundary preservation. PolMPCNN, HybridCVNet, NGDiff-SM, and SDNet generally recover the main land-cover regions, but still suffer from class leakage, over-smoothing, or loss of fine structures. SpectralDiff produces noticeable speckled errors, while F-Conv shows fragmented predictions in heterogeneous areas. In contrast, WGDNet generates cleaner homogeneous regions and more accurate boundaries. It achieves the best OA, AA, and Kappa values of (99.69\%), (98.80\%), and (99.51\%), respectively, and obtains the highest accuracies on four of the five classes while tying for the best result on \textit{Urban}, demonstrating superior and balanced classification performance.

\begin{figure}[t]
    \centering
    \includegraphics[width=\columnwidth]{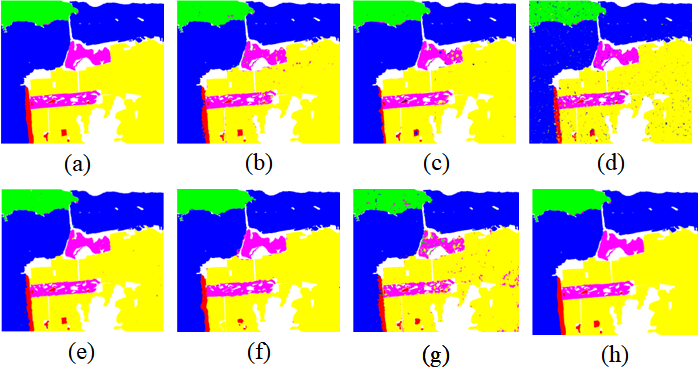}
    \caption{Classification maps on the San Francisco dataset.
    (a) Ground truth;
    (b) PolMPCNN;
    (c) HybridCVNet;
    (d) SpectralDiff;
    (e) NGDiff-SM;
    (f) SDNet;
    (g) F-Conv;
    and (h)  Proposed.}
    \label{fig5}
\end{figure}

\begin{table}[t]
\centering
\caption{Classification results (\%) of different methods on the  San Francisco dataset.}
\label{tab:san_francisco_results}
\resizebox{\columnwidth}{!}{
\begin{tabular}{lccccccc}
\hline
Class & PMC & HCV & SPD & NGD & SDN & FC & Ours \\
\hline
Bare soil  & 73.21&  52.19 &  76.06 & 90.33 & \underline{90.37} & 82.23 & \textbf{95.97} \\
Mountain   & 94.42 & \textbf{99.30} &  98.05 & \textbf{99.35} & 99.31 & 96.91 & 99.29 \\
Ocean      &  99.53 &  99.58 & 99.09 & 99.61 & \underline{99.77} & 99.43 & \textbf{99.87} \\
Urban      &  99.73 & 99.67 & 98.45 &  \textbf{99.85} & \underline{99.81} & 98.17 & \textbf{99.85} \\
Vegetation & 97.24 & 91.24 &  90.60 &  97.92 & \underline{97.95} & 78.19 & \textbf{99.00} \\
\hline
OA         &  97.93 &  98.31 &  97.49 & 99.42 & \underline{99.47} &  96.98 & \textbf{99.69} \\
AA         &  93.64 &  88.62 & 91.72 &  97.41 & \underline{97.44} & 90.98 & \textbf{98.80} \\
Kappa      &  98.64 &  97.33 & 96.04 &  99.09 & \underline{99.17} &95.26 & \textbf{99.51} \\
\hline
\end{tabular}
}
\end{table}

\subsubsection{Flevoland data set}The comparison methods show clear differences on the Flevoland dataset. PolMPCNN, HybridCVNet, SpectralDiff, NGDiff-SM, SDNet, and F-Conv generally preserve the main land-cover regions, but still suffer from class confusion, missed small structures, inaccurate boundaries, or scattered errors. In contrast, the proposed method produces more complete farmland parcels, cleaner forest and water regions, and more accurate urban boundaries. As reported in Table~\ref{tab:flevoland_results}, it achieves the best OA and Kappa of (99.82\%) and (99.80\%), respectively, together with the second-best AA of (99.43\%). It also obtains the best or tied-best results on eight classes, demonstrating superior overall accuracy, spatial consistency, and balanced discrimination across land-cover categories.

\begin{figure}[t]
    \centering
    \includegraphics[width=\columnwidth]{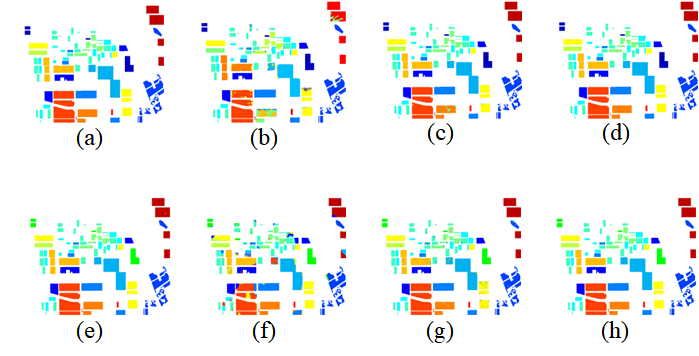}
    \caption{Classification maps on the San Francisco dataset.
    (a) Ground truth;
    (b) PolMPCNN;
    (c) HybridCVNet;
    (d) SpectralDiff;
    (e) NGDiff-SM;
    (f) SDNet;
    (g) F-Conv;
    and (h)  Proposed.}
    \label{fig5}
\end{figure}

\begin{table}[t]
\centering
\caption{Classification results (\%) of different methods on the Flevoland dataset.}
\label{tab:flevoland_results}
\scriptsize
\setlength{\tabcolsep}{2.2pt}
\renewcommand{\arraystretch}{0.92}

\resizebox{\columnwidth}{!}{
\begin{tabular}{lccccccc}
\hline
Class & PMC & HCV & SPD & NGD & SDN & FC & Ours \\
\hline
Stem beans & 99.78 & 99.52 & 99.69 & 99.69 & \underline{99.80} & \textbf{99.87} & 99.67 \\
Peas & \underline{99.98} & 99.26 & \textbf{100.00} & \textbf{100.00} & 99.65 & 99.30 & \textbf{100.00} \\
Forest & \underline{99.96} & 99.89 & 99.95 & 99.95 & \textbf{100.00} & 97.09 & \textbf{100.00} \\
Lucerne & 98.44 & \underline{99.92} & 96.20 & 96.20 & 99.89 & 97.61 & \textbf{99.96} \\
Wheat & 98.01 & 99.51 & \textbf{99.94} & \textbf{99.94} & \textbf{99.94} & 97.88 & \underline{99.93} \\
Beets & 95.51 & \textbf{100.00} & 99.20 & 99.20 & 99.08 & 99.22 & \underline{99.99} \\
Potatoes & 98.76 & 97.10 & \textbf{99.94} & \textbf{99.94} & 99.76 & 98.40 & \underline{99.81} \\
Bare soil & 95.70 & 97.86 & \textbf{100.00} & \textbf{100.00} & \textbf{100.00} & \underline{99.06} & 98.93 \\
Grasses & 99.83 & \underline{99.84} & 98.42 & 98.42 & 99.27 & 95.35 & \textbf{100.00} \\
Rapeseed & 93.55 & 88.54 & \textbf{100.00} & \textbf{100.00} & \underline{99.79} & 94.17 & 99.32 \\
Barley & 95.96 & 92.76 & \underline{99.75} & \underline{99.75} & \textbf{100.00} & 97.01 & \textbf{100.00} \\
Wheat2 & 40.24 & 99.28 & \textbf{100.00} & \textbf{100.00} & \underline{99.47} & 93.05 & \textbf{100.00} \\
Wheat3 & 93.88 & 99.19 & 99.92 & 99.92 & \underline{99.97} & 98.76 & \textbf{99.98} \\
Water & 96.01 & 99.84 & \underline{99.97} & \underline{99.97} & \textbf{100.00} & 99.95 & \textbf{100.00} \\
Buildings & 91.27 & 91.81 & 82.56 & 82.56 & \textbf{98.95} & \underline{95.11} & 93.76 \\
\hline
OA & 93.82 & 98.98 & \underline{99.73} & \underline{99.73} & 99.69 & 97.63 & \textbf{99.82} \\
AA & 93.06 & 98.35 & 98.57 & 98.57 & \textbf{99.64} & 97.45 & \underline{99.43} \\
Kappa & 93.26 & 98.88 & \underline{99.71} & \underline{99.71} & 99.66 & 97.41 & \textbf{99.80} \\
\hline
\end{tabular}
}
\end{table}

\subsection{Ablation Study}

The proposed WGDNet consists of the learnable Wishart convolution module and the GAnet module. To evaluate their contributions, four variants are compared: CNN, Fixed-Wishart-CNN(denoted by"FWC"), Learnable-Wishart-CNN(denoted by"LWC") and the proposed WGDNet. The OA, AA, Kappa, MIoU, and F-Score are used for evaluation. 

\begin{table}[ht]
\centering
\caption{ABLATION EXPERIMENT RESULTS ON  Xi'an Dataset ($\%$)}
\setlength{\tabcolsep}{1pt} 
\begin{tabular}{p{1.7cm}p{1.2cm}p{1.2cm}p{1.2cm}p{1.4cm}p{1.2cm}}
\hline
Method & OA & AA & Kappa & F1-score & MIoU  \\
\hline
CNN & 93.98 &70.48 & 90.17& 94.04 & 65.13 \\
FWC & 95.18 & 71.02 &92.08& 95.19 & 67.02 \\
LWC & 96.52 & 96.28 & 94.27& 96.54 &92.06 \\
WGDNet &97.94  & 97.73 &96.51  &95.43  &93.53 \\

\hline
\end{tabular}
\label{st3}
\end{table}

\begin{table}[ht]
\centering
\caption{ABLATION EXPERIMENT RESULTS ON  San Francisco  Dataset ($\%$)}
\setlength{\tabcolsep}{1pt} 
\begin{tabular}{p{1.7cm}p{1.2cm}p{1.2cm}p{1.2cm}p{1.4cm}p{1.2cm}}
\hline
Method & OA & AA & Kappa & F1-score & MIoU  \\
\hline
CNN & 91.10 &93.08 & 86.65& 92.04 &72.00 \\
FWC & 97.89 & 97.62 &96.27& 95.93 & 92.28 \\
LWC & 98.92 & 98.40 & 98.32& 98.95 &93.39 \\
WGDNet &99.69  & 98.80 &99.51  &99.69  &97.80 \\

\hline
\end{tabular}
\label{st3}
\end{table}

\begin{table}[ht]
\centering
\caption{ABLATION EXPERIMENT RESULTS ON  Flevoland   Dataset ($\%$)}
\setlength{\tabcolsep}{1pt} 
\begin{tabular}{p{1.7cm}p{1.2cm}p{1.2cm}p{1.2cm}p{1.4cm}p{1.2cm}}
\hline
Method & OA & AA & Kappa & F1-score & MIoU  \\
\hline
CNN & 92.90 &92.66 & 92.25& 92.84 &87.56 \\
FWC & 97.27 & 96.35 &95.95& 95.15 & 93.12 \\
LWC & 98.46 & 97.90 & 97.28& 97.15 &96.45 \\
WGDNet &99.82  & 99.43 &99.80  &99.82  &99.22 \\

\hline
\end{tabular}
\label{st3}
\end{table}

\begin{figure*}[!h]
	\centerline{\includegraphics[scale=0.55]{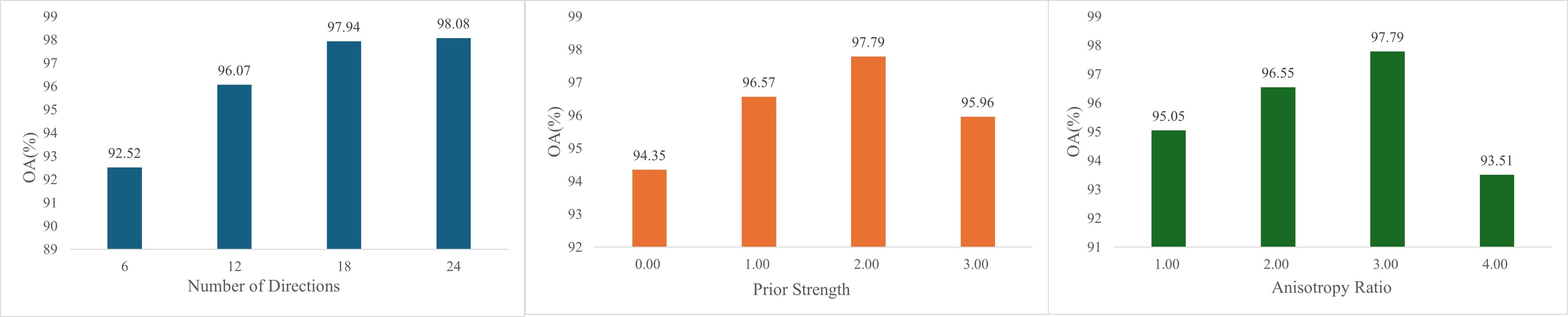}}
	\caption{The effect of different parameters on classification accuracies. (a) Effect of the number of directions. (b) Effect of prior strength. (c) Effect of the anisotropy ratio.}
	\label{3figures}
\end{figure*}

Table~\ref{st3} presents ablation results in the Xi’an, San Francisco, and Flevoland datasets. The CNN baseline yields poorer performance without scattering information. The fixed Wishart-based FWC outperforms CNN, although still lower than others. Further performance gains from LWC demonstrate its superiority with learned Wishart convolution. Our WGDNet, which integrates the learnable Wishart module with the GAnet, achieves optimal performance with OA values of 97.94\%, 99.69\%, and 99.82\% on the three datasets. It surpasses LWC by 1.42\%, 0.77\%, and 1.36\% in OA, verifying the benefit of adaptive geometric feature extraction. These results demonstrate the synergistic effect of the Wishart module and GAnet, which allows WGDNet to deliver an accurate and robust PolSAR classification.

\subsubsection{Effect of the Number of Directions}

Figs.~\ref{3figures}(a) illustrates that increasing the number of directions improves the directional representation capability of the model. When $D$ increases from 6 to 12 and 18, the OA improves from 92.52\% to 96.07\% and 97.94\%, respectively, demonstrating that sufficient directional sampling is beneficial for capturing irregular structures and complex boundaries. Although $D=24$ achieves a slightly higher OA of 98.08\%, it provides only a 0.14\% improvement over $D=18$ while requiring considerably more computation and processing time. Therefore, $D=18$ is selected with a compared performance and less computational efficiency.
\subsubsection{Effect of Prior Strength}

The prior strength controls the influence of the predicted orientation prior on directional feature aggregation. From Figs.~\ref{3figures}(b), it can be seen that When the prior strength increases from 0 to 1 and 2, the OA improves from 94.35\% to 96.57\% and 97.79\%, respectively. This demonstrates that appropriate orientation-prior guidance helps the model emphasize directionally consistent Wishart responses and better characterize irregular structures and object boundaries. However, when the prior strength increases to 3, the OA decreases to 95.96\%. This may be because an excessively strong prior suppresses the data-driven directional responses and reduces the adaptability of the aggregation process. Therefore, the prior strength is set to 2.

\subsubsection{Effect of the Anisotropy Ratio}

The anisotropy ratio controls the aspect ratio of the directional Wishart kernels and directly affects their sensitivity to oriented structures. Figs.~\ref{3figures}(c) shows that When the anisotropy ratio increases from 1 to 2 and 3, the OA improves from 95.05\% to 96.55\% and 97.79\%, respectively. This indicates that an appropriate anisotropic configuration enables the kernels to better capture elongated structures, directional boundaries, and local spatial variations. However, when the ratio increases further to 4, the OA decreases markedly to 93.51\%. An excessively large ratio produces overly elongated kernels, which may weaken local contextual modeling and reduce robustness to complex structures. Therefore, the anisotropy ratio is set to 3.

\subsection{Running Time Analysis}

Table~\ref{t-time} summarizes the training and testing efficiency of competing methods on the Xi’an dataset. The proposed method achieves a well-balanced tradeoff between performance and efficiency, with a training time of 383.14 s and a low inference latency of 2.43 s. It substantially reduces training overhead compared with PMC and SDN while retaining superior classification accuracy. Although HCV and FC train faster, their performance is constrained by limited modeling of polarimetric scattering characteristics. The introduced learnable Wishart statistical modeling and adaptive geometric feature extraction inevitably increase training complexity, yet the overhead remains tolerable. Meanwhile, the competitive testing time further validates the practical applicability of our method for PolSAR classification.

\begin{table}[ht]
\caption{Training and testing time of different methods on Xi'an dataset (s)}
\label{t-time}
\centering
\normalsize

\resizebox{\columnwidth}{!}{
\begin{tabular}{lccccccc}

\hline
Method & PMC & HCV & SPD & NGD & SDN & FC & Prop.\\
\hline
Train & 21201.26 & 99.84 & 731.89 & 71.73 & 4236.15 & 99.84 & 383.14\\
Test  & 1.96 & 9.19 & 5.16 & 8.26 & 2.45 & 4.29 & 2.43\\
\hline
\end{tabular}
}
\end{table}



\bibliography{aaai2026}

\end{document}